%% file: main.tex
\documentclass[10pt,twocolumn,letterpaper]{article}

\usepackage{iccv}              


\usepackage{microtype}      
\usepackage{xcolor}         
\usepackage{array}
\usepackage{adjustbox}
\usepackage{multirow}
\usepackage{hhline}
\usepackage{makecell}
\usepackage{textcomp}
\usepackage{booktabs}
\usepackage{amsmath}
\usepackage{amssymb}
\usepackage{graphicx}
\usepackage{subcaption}
\usepackage{afterpage}
\usepackage{placeins}
\usepackage{colortbl}
\usepackage{tabularx}
\usepackage{xcolor}
\usepackage{float}

\definecolor{iccvblue}{rgb}{0.21,0.49,0.74}
\usepackage[pagebackref,breaklinks,colorlinks,allcolors=iccvblue]{hyperref}

\def\paperID{11752} 
\def\confName{ICCV}
\def\confYear{2025}

\title{RCTDistill: Cross-Modal Knowledge Distillation Framework for Radar-Camera 3D Object Detection with Temporal Fusion}

\author{
Geonho Bang$^{1}$\thanks{Equal contributions}
\qquad
Minjae Seong$^{2}$\footnotemark[1]
\qquad
Jisong Kim$^{2}$\footnotemark[1]
\qquad
Geunju Baek$^{1}$
\\
Daye Oh$^{3}$
\qquad
Junhyung Kim$^{3}$
\qquad
Junho Koh$^{3}$
\qquad
Jun Won Choi$^{1}$\thanks{Corresponding author}
\and
$^{1}$  Seoul National University \quad $^{2}$Hanyang University \quad $^{3}$Hyundai Motor Company
\and
{\tt\small\{ghbang, gjbaek\}@spa.snu.ac.kr} 
\qquad {\tt\small \{mjseong, jskim\}@spa.hanyang.ac.kr} \\  
{\tt\small \{daye.oh, univjun, junhkoh\}@hyundai.com}  
\qquad{\tt\small junwchoi@snu.ac.kr}
}

\begin{document}
\maketitle
\input{sec/0_abstract}
\input{sec/1_intro}
\input{sec/2_related}
\input{sec/3_method}

\input{sec/4_experiment}
\input{sec/5_conclusion}
\input{sec/6_acknowledgement}

{
    \small
    \bibliographystyle{ieeenat_fullname}
    \bibliography{main}
}

\end{document}

%% file: sec/0_abstract.tex
\begin{abstract}
Radar-camera fusion methods have emerged as a cost-effective approach for 3D object detection but still lag behind LiDAR-based methods in performance. Recent works have focused on employing temporal fusion and Knowledge Distillation (KD) strategies to overcome these limitations. However, existing approaches have not sufficiently accounted for uncertainties arising from object motion or sensor-specific errors inherent in radar and camera modalities. In this work, we propose RCTDistill, a novel cross-modal KD method based on temporal fusion, comprising three key modules: Range-Azimuth Knowledge Distillation (RAKD), Temporal Knowledge Distillation (TKD), and Region-Decoupled Knowledge Distillation (RDKD). RAKD is designed to consider the inherent errors in the range and azimuth directions, enabling effective knowledge transfer from LiDAR features to refine inaccurate BEV representations. TKD mitigates temporal misalignment caused by dynamic objects by aligning historical radar-camera BEV features with current LiDAR representations. RDKD enhances feature discrimination by distilling relational knowledge from the teacher model, allowing the student to differentiate foreground and background features. RCTDistill achieves state-of-the-art radar–camera fusion performance on both the nuScenes and View-of-Delft (VoD) datasets, with the fastest inference speed of 26.2 FPS.
\end{abstract}

%% file: sec/1_intro.tex
\section{Introduction}
\label{sec:intro}

\begin{figure}[t]
    \centering
        \includegraphics[width=0.88\linewidth]{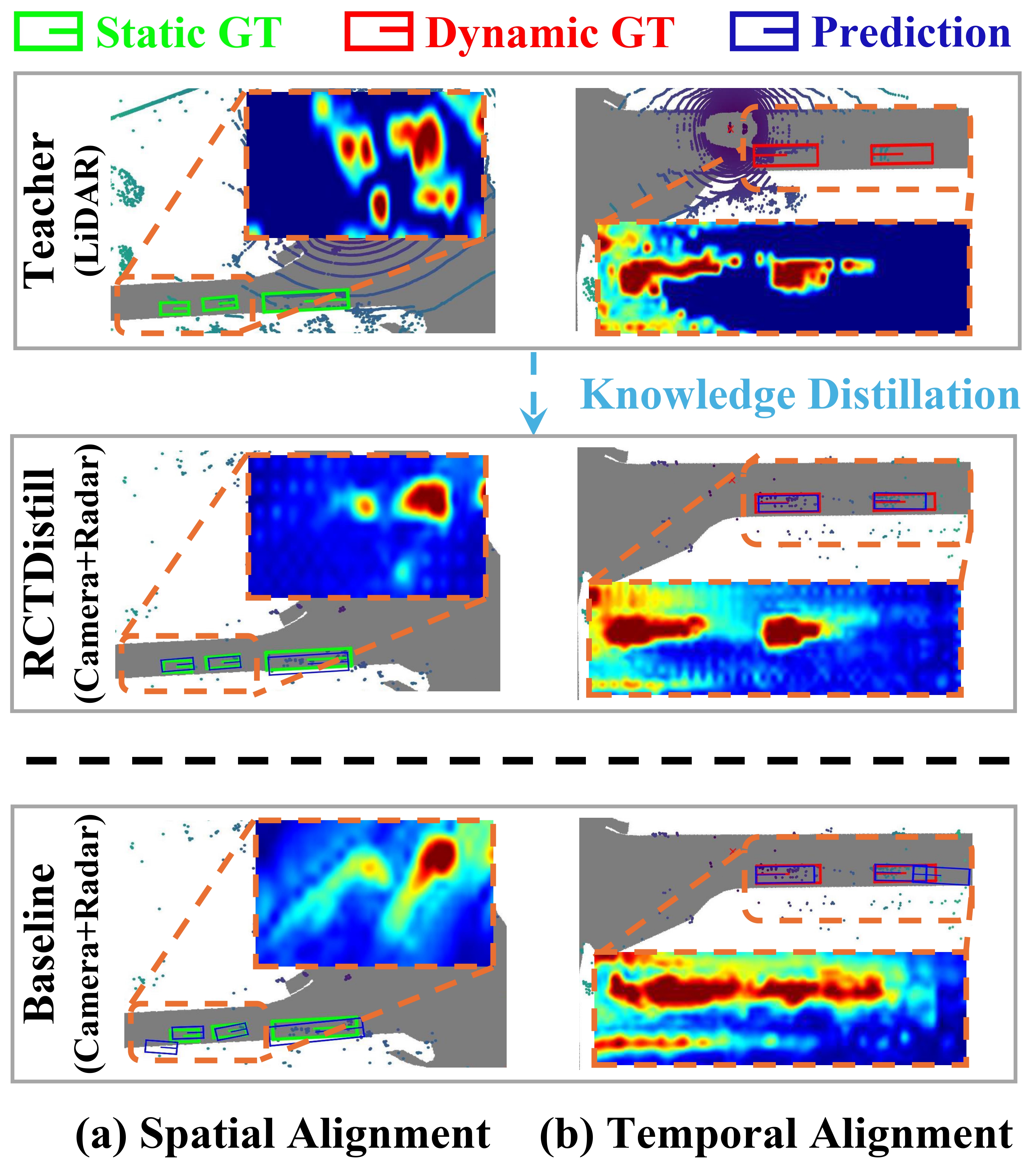}
        \caption{RCTDistill enhances the quality of BEV features for 3D object detection by aligning spatial features in the range-azimuth direction and temporal features in dynamic object trajectories.}
    \label{fig:Intro}
\end{figure}

Radar-camera-based 3D detection serves as a crucial component in autonomous driving systems.
While various Radar-Camera fusion approaches exist, recent methods predominantly focus on feature-level fusion, with particular emphasis on Bird's Eye View (BEV) domain \cite{rcm-fusion,crn,rcbevdet}. Although these methods effectively address the inherent limitations of individual sensors, there remains a notable performance gap when compared to LiDAR-based 3D detectors, as illustrated in Figure~\ref{fig:Intro}, where radar-camera fusion models show spatially and temporally misaligned BEV representations.

In an effort to overcome these performance gaps, Knowledge Distillation (KD) \cite{crkd, x3kd} and temporal fusion approaches \cite{crtfusion, crn, rctrans} have recently emerged as promising solutions in 3D object detection. KD-based approaches employ detectors based on low-fidelity sensors (e.g., cameras, radar, or camera-radar fusion) as student models while leveraging detectors based on higher-fidelity sensor inputs (e.g., LiDAR or LiDAR-camera fusion) as teacher models to enhance the student model's performance. In parallel, temporal fusion methods aim to improve performance by incorporating additional information from past frames and merging it with current features to compensate for information that may be missed at a single time point.

However, several significant challenges remain in these approaches. 
First, many existing KD methods do not fully account for sensor-specific characteristics \cite{x3kd, crkd}, suggesting that explicitly incorporating these characteristics could further improve KD effectiveness.
Cameras suffer from depth ambiguity, introducing uncertainties in object distance estimation during BEV feature generation \cite{gaussianlss, dme}, whereas radar provides reliable depth 
but has low angular resolution, leading to imprecise object localization in the horizontal plane \cite{sparc, zhou2023bridging}. While these sensors appear complementary in nature, integrating both modalities remains challenging, especially when it comes to mitigating their inherent uncertainties.
Second, despite recent advances \cite{crtfusion}, addressing error propagation caused by object movements remains a challenge for temporal fusion models.
The independent motion of dynamic objects leads to temporal misalignment, requiring explicit motion estimation to align features across frames. This process increases latency, ultimately degrading real-time detection performance.

In this paper, we propose RCTDistill, a novel cross-modal knowledge distillation method that transfers knowledge from a LiDAR model to a temporal radar-camera fusion model. RCTDistill performs knowledge distillation in three directions: Range-Azimuth Knowledge Distillation (RAKD), Temporal Knowledge Distillation (TKD), and Region-Decoupled Knowledge Distillation (RDKD). RAKD mitigates feature uncertainties in the fused BEV representation using an elliptical Gaussian mask, which accounts for varying uncertainties along the range and azimuth directions. By focusing on elliptical regions with Gaussian mask intensities above a threshold, it performs targeted feature distillation to effectively transfer knowledge from the teacher model. TKD addresses temporal feature misalignment in dynamic object detection by employing HA-Net to align historical BEV features. It generates trajectory-aware Gaussian regions for distillation, effectively capturing temporal dynamics while suppressing misaligned features. RDKD leverages an affinity map to distill relational knowledge between features, enabling the student model to learn discriminative feature relationships between foreground and background regions from the teacher model. RCTDistill achieves state-of-the-art performance and latency on both the nuScenes \cite{nuscenes} and View-of-Delft \cite{vod} radar-camera 3D detection benchmarks, outperforming all previous methods.

The main contributions of this paper are as follows:
\begin{itemize}
    \item Our RCTDistill is the first method to introduce a knowledge distillation technique specifically designed for radar-camera temporal fusion. Previous methods \cite{x3kd, crkd} focused solely on a single-frame setup, without explicitly utilizing temporal information. In this paper, we propose a novel cross-modal knowledge distillation approach that aligns BEV features over time, enhancing 3D object detection performance.
    \item We propose three knowledge distillation techniques, RAKD, TKD, and RDKD designed to address misalignment issues in radar-camera temporal 3D object detection. RAKD incorporates sensor-specific characteristics into the distillation process, while TKD focuses on temporal fusion by compensating for dynamic object motion. RDKD enhances the model's discriminative power by distilling the relationships between foreground and background features from the teacher model.
    \item RCTDistill achieves improvements of a 4.7\% mean Average Precision (mAP) and 4.9\% NuScenes Detection Score (NDS)  over the student model. Furthermore, it outperforms existing state-of-the-art radar-camera fusion methods on both the nuScenes and VoD datasets, while maintaining a real-time inference speed of 26.2 FPS.
\end{itemize}

%% file: sec/2_related.tex
\section{Related Work}
\label{sec:formatting}

\subsection{Radar-Camera 3D Object Detection}
Recent radar-camera fusion models have adopted view transformation-based methods, which can be categorized into two directions: implicitly mapping features through attention mechanisms \cite{detr3d, petr, bevformer}, or explicitly using depth distributions obtained from camera features \cite{lss, bevdepth}. RCM-Fusion \cite{rcm-fusion}, following the former approach, integrated radar and camera features into a unified bird's-eye-view (BEV) space using deformable attention. Similarly, SpaRC \cite{sparc} enhances the positional information of sparse object queries and 2D camera features through an attention mechanism based on radar point features. Conversely, CRN \cite{crn} and CRT-Fusion \cite{crtfusion} adopted the latter strategy to enhance camera BEV features via radar-guided view transformation before feature-level fusion. RCBEVDet \cite{rcbevdet} introduced a dual-path radar backbone combining point-based and transformer-based modules, leveraging deformable cross-attention for effective radar-camera feature fusion. 
To incorporate temporal modeling, some works \cite{crn, rcbevdet, crtfusion, sparc} have extended their methods by integrating multi-frame radar and camera features.
Notably, CRT-Fusion \cite{crtfusion} introduces dynamic object modeling to mitigate temporal misalignment.

\begin{figure*}[t]
    \centering
        \includegraphics[width=0.88\linewidth]{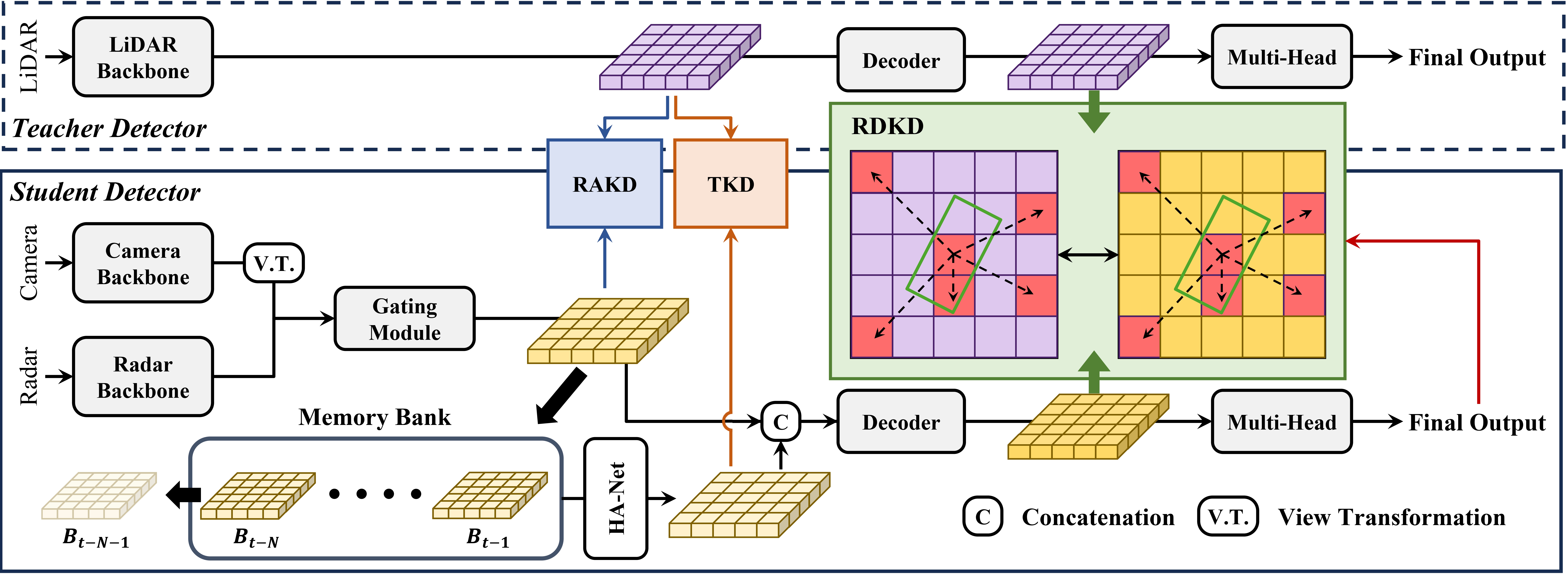}
        \caption{\textbf{Overall architecture of RCTDistill.} The student model fuses radar and camera features using a gating module to generate low-level BEV features, which are stored in a memory bank. These historical features are then temporally aggregated with the current BEV features, followed by a decoder that produces high-level BEV representations. RAKD and TKD enhance the low-level features by accounting for sensor uncertainty and object motion, respectively, while RDKD learns discriminative feature relationships between foreground and background regions from the teacher detector. Note that the teacher model is utilized only during training and omitted during inference.}
    \label{fig:overall}
\end{figure*}

\subsection{Knowledge Distillation in 3D Object Detection}
Cross-modal KD approaches have been widely adopted to facilitate knowledge transfer across heterogeneous sensor modalities to overcome sensing limitations \cite{ligastereo, monodistill, cmkd, uvtr, s2m2ssd, bevdistill, unidistill, x3kd, distillbev}.
BEVDistill \cite{bevdistill} proposed feature- and instance-level distillation in the BEV space, and UniDistill \cite{unidistill} employed flexible KD pathways—such as LiDAR-to-camera, camera-to-LiDAR, and fusion-to-single modality—by projecting both teacher and student features into a unified BEV representation.

Growing interest in leveraging radar for 3D object detection has led to increased attention on cross-modal KD, where data sparsity remains a key challenge.
RadarDistill \cite{bang2024radardistill} and CRKD \cite{crkd} addressed this by transferring rich knowledge from LiDAR or LiDAR-camera models using feature- and relation-level supervision.
LEROjD \cite{lerojd} and SCKD \cite{sckd} further explored 4D radar-based KD; LEROjD employed a multi-stage training and distillation framework to leverage the spatial density of LiDAR, while SCKD adopted a semi-supervised approach using a LiDAR-camera teacher.

In parallel, temporal KD approaches have emerged to address object misalignment across time, a critical issue in dynamic driving scenes.
VCD \cite{vcd} projected the historical positions of ground truth into the current frame and applied KD at these aligned positions. STXD \cite{stxd} leveraged similarity maps to align past frames of the teacher model with both the current frame of the teacher and the student.

%% file: sec/3_method.tex
\section{Method}
\label{main_sec:method}
The overall architecture of RCTDistill is illustrated in Figure~\ref{fig:overall}. Our model leverages a cross-modal knowledge distillation framework to enhance the performance of radar-camera fusion models. In Section~\ref{sec3.1:preliminaries}, we provide a comprehensive overview of our teacher and student model architectures. Sections~\ref{sec:RAKD}--\ref{sec:RDKD} present the detailed descriptions of our proposed knowledge distillation methods, RAKD, TKD, and RDKD.

\begin{figure*}[t]
    \centering
        \includegraphics[width=0.88\linewidth]{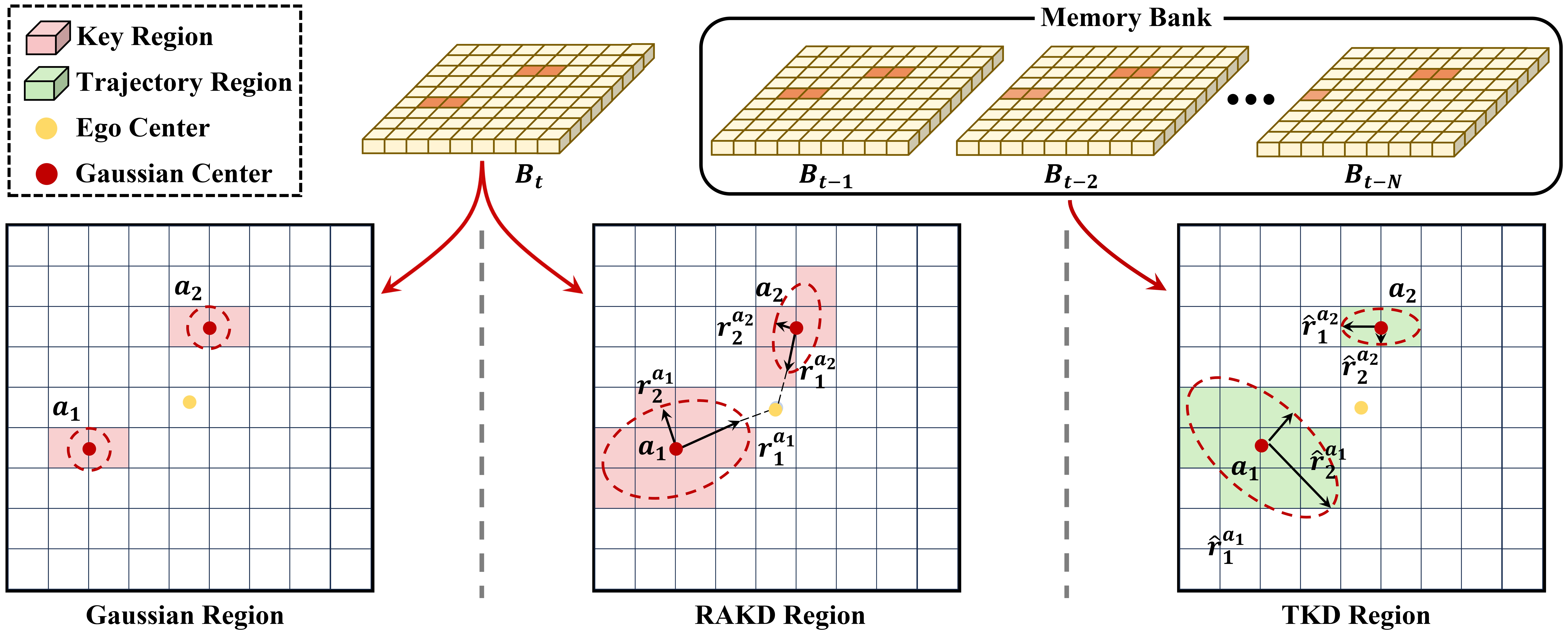 }
        \caption{\textbf{Elliptical Gaussian Mask Regions for RAKD and TKD.} Elliptical masks in the RAKD (center) and TKD (right) regions are used for knowledge distillation, targeting range-azimuth uncertainty and temporal misalignment, respectively.} 
    \label{fig:KD_methods}
\end{figure*}

\subsection{Baseline Model}
\label{sec3.1:preliminaries}
\noindent{\bf Student model.} 
We adopt BEVFusion-R as our student model, which extends the BEVFusion \cite{bevfusion} framework for radar-camera fusion. The model extracts radar BEV features and camera features using separate backbones. These camera features are then transformed into BEV representation via a View Transformation (VT) module. Subsequently, the radar and camera BEV features are concatenated and fused through a $1 \times 1$ convolution, followed by 2D CNN blocks and a detection head to produce the final predictions.

We improve the performance of the baseline by incorporating several modifications to BEVFusion-R. First, for multi-modal feature fusion, we replace $1 \times 1$ convolution layer with an adaptive gating network \cite{crkd} that adaptively adjusts the influence of each modality. This gating network outputs low-level fused BEV features  $B^{\text{RC}}_{\text{low}} \in \mathbb{R}^{D \times H \times W}$.  These BEV features are aggregated using a streaming-based temporal fusion, where a memory bank retains  historical BEV features obtained from previous time steps. These historical features are transformed into the current coordinate and then concatenated channel-wise with the current BEV features. Finally, the temporally fused features pass through a decoder consisting of 2D CNN layers to produce high-level  BEV features $B^{\text{RC}}_{\text{high}} \in \mathbb{R}^{D' \times H \times W}$.

\noindent{\bf Teacher model.} 
We employed CenterPoint, a widely adopted LiDAR-based 3D object detector, as our teacher model. CenterPoint groups raw LiDAR point clouds into 3D voxels, then extracts low-level BEV features \( B^{L}_{\text{low}} \in \mathbb{R}^{C \times H \times W} \) through a 3D sparse convolutional backbone. These low-level features are further processed by a decoder consisting of 2D CNN layers to generate high-level BEV features, \( B^{L}_{\text{high}} \in \mathbb{R}^{C' \times H \times W} \). 

\subsection{Range-Azimuth Knowledge Distillation}
\label{sec:RAKD}
Cross-modal knowledge distillation aims to  transfer the desirable representations from the teacher modality to the student modality. Earlier methods \cite{bevdistill, simdistill} have used ground truth (GT) box-centered Gaussian masks to determine the region for feature distillation, as shown in Figure~\ref{fig:KD_methods} (left). 
However, these approaches fail to capture modality-specific characteristics, such as depth ambiguity in cameras and range-azimuth uncertainties in radar. Gaussian masks apply weights that decrease uniformly from the object center, which makes it difficult to adapt to the unique distribution of each modality. To address these challenges, we propose Range-Azimuth Knowledge Distillation (RAKD), which utilizes an elliptical Gaussian mask. 
The shape of the ellipse is determined by the inherent uncertainties along range and azimuth directions, as shown in Figure~\ref{fig:KD_methods} (middle).

For each $i$-th GT 3D box projected to bird's eye view, we obtain a 2D box defined by center position $(p_x^i, p_y^i)$, heading angle $\theta^i$, and dimensions $(l^i, w^i)$. Inspired by DORN \cite{DORN}, we determine the radius $(r_1^i, r_2^i)$ of an elliptical region in the major and minor axes  based on the object size and distance from the ego-vehicle as
\begin{equation}
r_1^i = l^i\cdot \left( \alpha_l \over l^i \right)^{\beta}, \quad r_2^i = w^i\cdot \left( \alpha_w \over w^i \right)^{\beta},
\end{equation}
where $\alpha_l$ and $\alpha_w$ denote the hyperparameters that control the scale of $r_1^i$ and $r_2^i$, and $\beta (<1)$ denotes the normalized distance from the ego-vehicle to the center position of the object. Notice that $r_1^i$ and $r_2^i$ are proportional to the size $l^i$ and $w^i$ of the 2D box, respectively.  Using these parameters, we generate an Elliptical Gaussian mask $E_{i,x,y}$ of the $i$-th GT box at the position $(x, y)$ as
\begin{equation}
\label{eq:elliptical_gaussian}
E_{i, x, y} = \exp\left(-\frac{1}{2} \left(\frac{(x')^2}{(r_1^i)^2} + \frac{(y')^2}{(r_2^i)^2}\right)\right),
\end{equation}
where
\begin{equation}
\begin{bmatrix}
x' \\ 
y' 
\end{bmatrix} = 
\begin{bmatrix}
\cos \theta^i & -\sin \theta^i \\ 
\sin \theta^i & \cos \theta^i 
\end{bmatrix}
\begin{bmatrix}
x - p_x^i \\ 
y - p_y^i 
\end{bmatrix}.
\end{equation}
We then merge the elliptical Gaussian masks \( E \in \mathbb{R}^{N \times H \times W} \), generated from the $N$ GT boxes, into \( \bar{E} \in \mathbb{R}^{H \times W} \). 
When elliptical masks overlap between objects, $\bar{E}_{x,y}$ is obtained by taking the maximum magnitude in those areas.
Next, we obtain a Distillation Mask \( W_\text{RA} \in \mathbb{R}^{H \times W} \) by assigning the corresponding value from $\bar{E}_{x,y}$ to each spatial position $(x,y)$  if it surpasses the threshold $\tau$, and zero otherwise. RAKD aims to align the low-level features of the student model with those of the teacher model within the Distillation Mask. To this end, we introduce a RAKD loss function
\begin{equation}
L_\text{RA} = \frac{1}{|N_\text{RA}|} \sum^{H}_{j=1}\sum^{W}_{k=1} W_{\text{RA}, j,k} \left\| B^{\text{L}}_{\text{low},j,k} - \bar{B}^{\text{RC}}_{\text{low},j,k} \right\|_{2},
\end{equation}
where $N_\text{RA}$ represents the number of non-zero elements in $W_\text{RA}$. The student feature map \(\bar{B}^{\text{RC}}_{\text{low}}\) is obtained by applying a 
\(1 \times 1\) convolution to the original features \(B^{\text{RC}}_{\text{low}}\) to match its channel dimension with that of $B^{\text{L}}_{\text{low},i,j}$.

\subsection{Temporal Knowledge Distillation}
\label{sec:TKD}
While previous temporal fusion methods \cite{crn, rcbevdet} in 3D object detection have achieved significant performance improvements through simple concatenation of historical BEV features followed by CNN layers, they often struggle to adequately capture the movement of dynamic objects. To address this issue, we propose Temporal Knowledge Distillation (TKD), a novel framework that effectively takes object motion information into account for knowledge distillation.

As illustrated in Figure~\ref{fig:overall}, we first concatenate historical BEV feature maps and process them through Historical Alignment Network (HA-Net), a 2D CNN-based architecture. In the presence of rapid object motion, BEV features exhibit significant variations between consecutive frames.  HA-Net aggregates these BEV features through large-scale kernel CNNs to alleviate inter-frame variations. The detailed architecture of HA-Net is provided in the Supplementary Material.

\begin{figure}[t]
    \centering
        \includegraphics[width=0.9\linewidth]{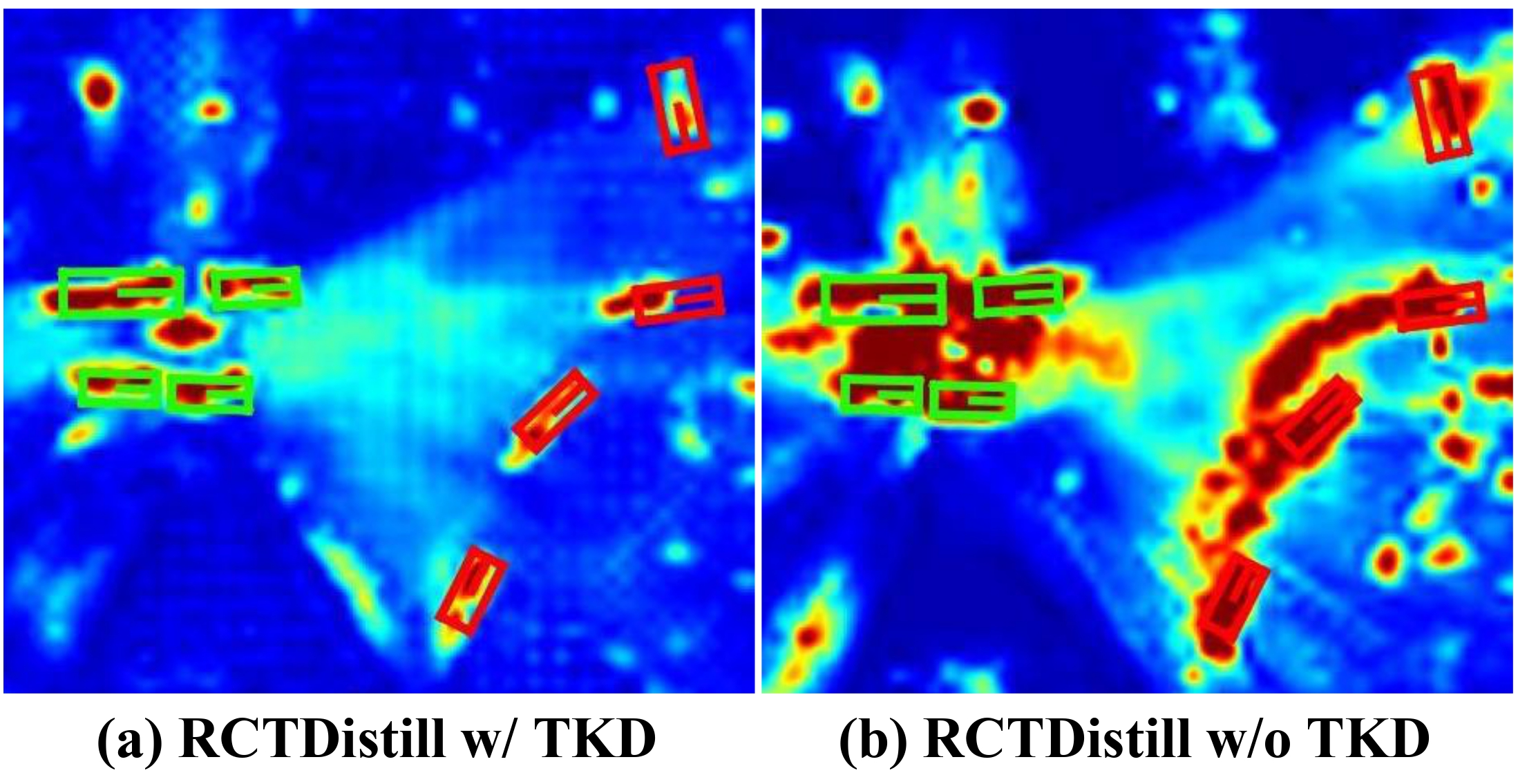 }
    \caption{\textbf{Visualization of the BEV feature map.} Green boxes denote static object Ground Truth (GT) boxes, Red boxes indicate dynamic object GT boxes.}
    \label{fig:tkd}
\end{figure}

We identify the spatial regions in temporal features for knowledge transfer.
To achieve this, we introduce temporal elliptical Gaussian masks $T \in \mathbb{R}^{N \times H \times W}$ whose shapes are determined based on object trajectories as shown in Figure~\ref{fig:KD_methods} (right). These masks are designed to transfer temporal representations within regions that encompass the trajectories of dynamic objects.
For the $i$-th object, we determine the elliptical Gaussian mask $T_{i}$  by revising the center position and major axis radius of the elliptical function $E_{i}$ in Equation~\ref{eq:elliptical_gaussian}. Given the velocity vector $\mathbf{v}^i=(v^i_x, v^i_y)$ and the center point $\mathbf{p}^i= (p^i_x, p^i_y)$ for the $i$-th object, the ellipse center $\mathbf{\hat{p}}^i=(\hat{p}^i_x, \hat{p}^i_y)$ is computed as
\begin{equation}
\mathbf{\hat{p}}^i = 
\begin{cases} 
\mathbf{p}^i - \frac{t_s}{2}\mathbf{v}^i, & \text{if } \|\mathbf{v}^i\|_2^2 > \tau_{v} \\
\end{cases}
\end{equation}
where $t_s$ is the temporal duration chosen to be less than the total duration of the historical frame. This implies that when the magnitude of the velocity vector exceeds a threshold  $\tau_{v}$, the center of the ellipse is obtained by shifting the object center in the direction opposite to its motion.   Subsequently,  the  radius \( \hat{r}^{i}_1 \) and \( \hat{r}^{i}_2 \) in the major and minor axis is calculated as
\begin{align}
\hat{r}^{i}_1 &= l^{i} + \sqrt{(\hat{p}^i_x - p^i_x)^2 + (\hat{p}^i_y - p^i_y)^2} \\
\hat{r}^{i}_2 &= w^{i},
\end{align}
to allow the masks to cover a historical trajectory of the $i$-th dynamic object.
Using the computed \( \mathbf{\hat{p}}^i \), \( \hat{r}^{i}_1 \) and \( \hat{r}^{i}_2 \), the temporal mask $T_{i,x,y}$ is given by
\begin{equation}
T_{i,x,y} = \exp\left(-\frac{1}{2} \left(\frac{(x'')^2}{(\hat{r}_1^i)^2} + \frac{(y'')^2}{(\hat{r}_2^i)^2}\right)\right),
\end{equation}
where
\begin{equation}
\begin{bmatrix}
x'' \\
y''
\end{bmatrix} =
\begin{bmatrix}
\cos \theta^i & -\sin \theta^i \\
\sin \theta^i & \cos \theta^i
\end{bmatrix}
\begin{bmatrix}
x - \hat{p}^i_x \\
y - \hat{p}^i_y
\end{bmatrix}.
\end{equation}
When the elliptical masks overlap, we obtain \( \bar{T} \in \mathbb{R}^{H \times W} \) by taking the maximum value of \( T_{i,x,y} \) over \( i \). Similarly to RAKD, the Temporal Distillation Mask \( W_{\text{T}} \in \mathbb{R}^{H \times W} \) is computed by retaining \( \bar{T}_{x,y} \) at position \((x, y)\) if it exceeds the threshold \( \tau \), or setting it to zero otherwise.
To facilitate temporal knowledge transfer in these regions, we 
introduce a temporal distillation loss function 
\begin{equation}
L_{\text{T}} = \frac{1}{|N_{\text{T}}|} \sum_{j=1}^{H} \sum_{k=1}^{W} W_{\text{T},j,k} \left\| B^{L}_{\text{low},j,k}-\hat{B}^{RC}_{\text{low},j,k} \right\|^2,
\end{equation}
where \( N_{\text{T}} \) denotes the number of non-zero elements in \( W_{\text{T}} \),  \( \hat{B}^{RC}_{\text{low}} \) represents the temporally aggregated feature maps obtained through  the HA-Net, and \( B^{L}_{\text{low}} \) denotes the current low-level feature map from the teacher model. This distillation process aims to align the historical camera-radar features with the current LiDAR features within the regions defined by $W_{\text{T}}$. As shown in Figure~\ref{fig:tkd}, the BEV feature maps produced with our proposed TKD module exhibit clearer object boundaries and reduced temporal artifacts for dynamic objects. In contrast, the baseline displays noticeable trailing effects around moving objects. The enhanced temporal alignment achieved by TKD leads to more accurate localization of dynamic objects at the current timestamp.

\subsection{Region-Decoupled Knowledge Distillation}
\label{sec:RDKD}
High-level feature maps—especially those from LiDAR-based models—are effective at capturing rich semantic information that distinctly separates foreground objects from background regions. To transfer this discriminative capability to radar-camera models, we propose RDKD, which aligns the relational structure between foreground and background regions in the high-level feature maps. Guided by the teacher model, the student model is encouraged to maintain high similarity among foreground features while effectively distinguishing them from background features.

\input{table/main_0307_jisong}

The high-level feature map $B^{\text{RC}}_{\text{high}}$ obtained from the student detector is processed through the detection head to generate a classification score map $B^{\text{RC}}_{\text{cls}}$. We obtain the confidence score map $\bar{B}^{RC}_{cls} \in \mathbb{R}^{H \times W}$ by taking the maximum value of $B^{\text{RC}}_{\text{cls}}$ across all classes. From $\bar{B}^{\text{RC}}_{\text{cls}}$, we identify all positions with confidence scores above a predefined threshold $\tau$. At these $K$ selected positions, we extract the corresponding high-level features from $\bar{B}^{\text{RC}}_{\text{cls}}$, resulting in a set of $K$ feature vectors, denoted as ${\mathbf{f}_1^{\text{RC}}, \mathbf{f}_2^{\text{RC}}, \dots, \mathbf{f}_K^{\text{RC}}}$. Using these extracted features, the $K \times K$ affinity map is computed as 
\begin{equation}
\label{eq:affinity_map}
S^{\text{RC}}_{jk} = \frac{\mathbf{f}_j^{\text{RC}} \cdot \mathbf{f}_k^{\text{RC}}}{\|\mathbf{f}_j^{\text{RC}}\|_2 \|\mathbf{f}_k^{\text{RC}}\|_2}, \quad \text{for } j, k = 1, 2, \ldots, K,
\end{equation}
where $\cdot$ denotes the inner product operation. 
From the same $K$ selected positions, we also extract $K$ high-level feature vectors from the high-level feature map $B^{\text{L}}_{\text{high}}$ obtained by the teacher model. 
Then, we compute the teacher affinity map $S^{\text{L}}$ in the same manner as Equation~\ref{eq:affinity_map}.

We define the RDKD loss $L_{RD}$ to guide the student model to learn the ability to discriminate between the foreground and background regions from the teacher model as
\begin{equation}
L_\text{RD} = \frac{1}{|K^2|} \sum^{K}_{j=1}\sum^{K}_{k=1} \left| S^{L}_{jk} - S^{RC}_{jk} \right|.
\end{equation}

\subsection{Loss Function}
The total loss function comprises a 3D detection loss $L_{det}$ and three KD loss terms. The total loss $ L_{total}$ is obtained as
\begin{equation}
L_{total} = L_{det} + \lambda_{\text{RA}} L_{\text{RA}} + \lambda_{\text{T}} L_{\text{T}} + \lambda_{\text{RD}} L_{\text{RD}},
\end{equation}
where $\lambda_{\text{RA}}$, $\lambda_{\text{T}}$, and $\lambda_{\text{RD}}$ are hyperparameters that control the relative importance of each loss term.

%% file: table/main_0307_jisong.tex
\newcolumntype{C}{>{\centering\arraybackslash}p{2.3em}}
\newcolumntype{S}{>{\centering\arraybackslash}p{2.7em}}
\newcolumntype{'}{!{\vrule width 0.1pt}}
\renewcommand{\arraystretch}{1.0}

\begin{table*}[t]

\begin{center}

\begin{adjustbox}{width=1.0\linewidth}
\definecolor{lightblue}{RGB}{220, 230, 255}
{
\fontsize{20pt}{25pt}\selectfont
\begin{tabular}{c ' c ' c ' c ' c ' c ' c ' C  C ' S S S S c ' c}
\Xhline{9\arrayrulewidth} 
Dataset & Methods  & Input & KD & Backbone & Image Size & Frames & NDS & mAP & mATE & mASE & mAOE & mAVE & mAAE & FPS\\ 
\Xhline{7\arrayrulewidth} 
\multirow{21}{*}{Val.} & Teacher \cite{centerpoint} & L & - & - & - & 1 &  67.1 & 60.3 & 0.298 & 0.251 & 0.288 & 0.281 & 0.190 & - \\
\cline{2-15}
 & BEVDepth$\textsuperscript{\textdagger}$ \cite{bevdepth} & C & - &  R50 & 256$\times$704 & 2 &    47.5 & 35.1 & 0.639 & 0.267 & 0.479 & 0.428 & 0.198 & 11.6 \\
 & SOLOFusion$\textsuperscript{\textdagger}$ \cite{solofusion} & C &  - & R50 & 256$\times$704 & 16+1 &    53.4 & 42.7 & 0.567 & 0.274 & 0.411 & 0.252 & 0.188 & 11.4 \\
 \cline{2-15}
  & X3KD$\textsuperscript{\textdagger}$ \cite{x3kd} & C+R & \checkmark &  R50 & 256$\times$704 & 1 &     53.8 & 42.3 & 0.487 & 0.277 & 0.542 & 0.344 & 0.197 & - \\
  & CRKD$\textsuperscript{\textdagger}$ \cite{crkd} & C+R & \checkmark &  R50 & 256$\times$704 & 1 &     54.9 & 43.2 & 0.450 & 0.267 & 0.442 & 0.339 & 0.176 & - \\
  & \cellcolor{lightblue}RCTDistill-S$\textsuperscript{\textdagger}$&  \cellcolor{lightblue}C+R &  
 \cellcolor{lightblue}\checkmark&
 \cellcolor{lightblue}R50 & \cellcolor{lightblue}256$\times$704 & \cellcolor{lightblue}1 &    \cellcolor{lightblue}\bf{55.5} & \cellcolor{lightblue}\bf{46.4} & \cellcolor{lightblue}0.488 & \cellcolor{lightblue}0.262 & \cellcolor{lightblue}0.462 & \cellcolor{lightblue}0.381 & \cellcolor{lightblue}0.176 & \cellcolor{lightblue}\bf{28.0} \\
  \cline{2-15}
 & CRN \cite{crn} & C+R & - &  R50 & 256$\times$704 & 4 &     56.0 & 49.0 & 0.487 & 0.277 & 0.542 & 0.344 & 0.197 & 20.4 \\
 & RCBEVDet$\textsuperscript{\textdagger}$ \cite{rcbevdet} &  C+R & - &  R50 & 256$\times$704 & 2 &    56.8 & 45.3 & 0.486 & 0.285 & 0.404 & 0.220 & 0.192 & 21.3 \\
 & CRT-Fusion$\textsuperscript{\textdagger}$ \cite{crtfusion} &  C+R & - &  R50 & 256$\times$704 & 6+1 &    59.7 & 50.8 & 0.461 & 0.264 & 0.419 & 0.234 & 0.186 & 14.5 \\
  & RCTrans$\textsuperscript{\textdagger}$ \cite{rctrans} &  C+R & - &  R50 & 256$\times$704 & 4 & 58.6 & 50.9 & 0.537 & 0.269 & 0.491 & 0.203 & 0.183 & 19.2 \\
 & SpaRC \cite{sparc} &  C+R & - &  R50 & 256$\times$704 & 8 &    62.0 & 54.5 & 0.496 & 0.269 & 0.403 & 0.177 & 0.181 & 19.1 \\
  & \cellcolor{gray!20}Baseline & \cellcolor{gray!20}C+R & \cellcolor{gray!20}- & \cellcolor{gray!20}R50 & \cellcolor{gray!20}256$\times$704 & \cellcolor{gray!20}8+1 &  \cellcolor{gray!20}57.3 & \cellcolor{gray!20}50.5 & \cellcolor{gray!20}0.491 & \cellcolor{gray!20}0.271 & \cellcolor{gray!20}0.560 & \cellcolor{gray!20}0.280 & \cellcolor{gray!20}0.189 & \cellcolor{gray!20}27.8  \\
 & \cellcolor{lightblue}RCTDistill &  \cellcolor{lightblue}C+R &  
 \cellcolor{lightblue}\checkmark&
 \cellcolor{lightblue}R50 & \cellcolor{lightblue}256$\times$704 & \cellcolor{lightblue}8+1 &    \cellcolor{lightblue}\bf{62.2} & \cellcolor{lightblue}\bf{55.2} & \cellcolor{lightblue}0.434 & \cellcolor{lightblue}0.241 & \cellcolor{lightblue}0.432 & \cellcolor{lightblue}0.263 & \cellcolor{lightblue}0.174 & \cellcolor{lightblue}\bf{26.2} \\
\Xcline{2-15}{2.5pt}
 & Teacher \cite{centerpoint} & L & - & - & - & 1 &  67.3 & 60.6 & 0.286 & 0.251 & 0.315 & 0.261 & 0.185 & - \\
 \cline{2-15}
 & BEVDepth$\textsuperscript{\textdagger}$ \cite{bevdepth} & C & - & R101 & 512$\times$1408 & 2 &     53.5 & 41.2 & 0.565 & 0.266 & 0.358 & 0.331 & 0.190 & 5.0 \\
 & SOLOFusion$\textsuperscript{\textdagger}$ \cite{solofusion} & C & - & R101 & 512$\times$1408 & 16+1 &     58.2 & 48.3 & 0.503 & 0.264 & 0.381 & 0.246 & 0.207 & 11.4 \\
 \cline{2-15}
 & CRN \cite{crn} & C+R & - & R101 & 512$\times$1408 & 4 &     59.2 & 52.5 & 0.460 & 0.273 & 0.443 & 0.352 & 0.180 & 7.2 \\
& CRT-Fusion \cite{crtfusion} &  C+R & - &   R101 & 512$\times$1408 & 6+1 &     62.1 & 55.4 & 0.425 & 0.264 & 0.433 & 0.237 & 0.193 & 4.9 \\
& SpaRC \cite{sparc} &  C+R & - & R101 & 512$\times$1408 & 8 &    64.4 & 57.1 & 0.484 & 0.264 & 0.308 & 0.175 & 0.178 & 7.2 \\  
& \cellcolor{gray!20}Baseline & \cellcolor{gray!20}C+R & \cellcolor{gray!20}- 
& \cellcolor{gray!20}R101 & \cellcolor{gray!20}512$\times$1408
& \cellcolor{gray!20}8+1 &  \cellcolor{gray!20}60.9 
& \cellcolor{gray!20}54.2 & \cellcolor{gray!20}0.433 
& \cellcolor{gray!20}0.272 & \cellcolor{gray!20}0.469 
& \cellcolor{gray!20}0.252 & \cellcolor{gray!20}0.198 &\cellcolor{gray!20} 8.7 \\ 
 & \cellcolor{lightblue}RCTDistill &  \cellcolor{lightblue}C+R &  \cellcolor{lightblue}\checkmark&
 \cellcolor{lightblue}R101 & \cellcolor{lightblue}512$\times$1408 & \cellcolor{lightblue}8+1 &     \cellcolor{lightblue} \bf{65.5} & \cellcolor{lightblue}\bf{59.0} & \cellcolor{lightblue}0.378 & \cellcolor{lightblue}0.250 & \cellcolor{lightblue}0.346 & \cellcolor{lightblue}0.236 & \cellcolor{lightblue}0.186 & \cellcolor{lightblue}\bf{8.4} \\
\Xhline{7\arrayrulewidth} 
 \multirow{8}{*}{Test} & BEVDepth \cite{bevdepth} &  C & - & ConvX-B & 640$\times$1600 & 2 &     60.9 & 52.0 & 0.445 & 0.243 & 0.352 & 0.347 & 0.128 & - \\
 & SOLOFusion \cite{solofusion} &  C & - & ConvX-B & 640$\times$1600 & 16+1 & 61.9 & 54.0 & 0.453 & 0.257 & 0.376 & 0.276 & 0.148  & - \\
 \cline{2-15}
 & CRN$\textsuperscript{\textdaggerdbl}$ \cite{crn} &  C+R & - & ConvX-B & 640$\times$1600 & 4 &  62.4 & 57.5 & 0.416 & 0.264 & 0.456 & 0.365 & 0.130 & - \\
 & RCBEVDet \cite{rcbevdet} &  C+R & - & V2-99 & 640$\times$1600 & 2 & 63.9 & 55.0 & 0.390 & 0.234 & 0.362 & 0.259 & 0.113 & - \\
 & CRTFusion \cite{crtfusion} &  C+R & - & ConvX-B & 512$\times$1408 & 6+1 & 64.9 & 58.3 & 0.365 & 0.261 & 0.405 & 0.262 & 0.132 & 3.7 \\
  & RCTrans \cite{rctrans} &  C+R & - &  V2-99 & 640$\times$1600 & 4 & 64.7 & 57.8 & 0.459 & 0.245 & 0.392 & 0.198 & 0.121 & - \\
 & SpaRC \cite{sparc} &  C+R & - & V2-99 & 640$\times$1600 & 8 &    67.1 & 60.0 & - & - & - & - & - & - \\

 & \cellcolor{lightblue}RCTDistill &  \cellcolor{lightblue}C+R & 
 \cellcolor{lightblue}\checkmark&
 \cellcolor{lightblue}ConvX-B & \cellcolor{lightblue}512$\times$1408 & \cellcolor{lightblue}8+1 &   \cellcolor{lightblue} \bf{67.4} & \cellcolor{lightblue} \bf{60.3} & \cellcolor{lightblue}0.334 & \cellcolor{lightblue}0.245 & \cellcolor{lightblue}0.334 & \cellcolor{lightblue}0.251 & \cellcolor{lightblue}0.113 & \cellcolor{lightblue}\bf{5.0} \\
\Xhline{9\arrayrulewidth} 

\end{tabular}
}
\end{adjustbox}

\caption{Performance comparisons with 3D object detectors on the nuScenes validation (Val.) and test set. Teacher denotes CenterPoint \cite{centerpoint}. `C', and `R' represent camera and radar, respectively. RCTDistill-S indicates the use of single-frame input. $\dagger$: trained with CBGS \cite{cbgs}. $\ddagger$: use Test Time Augmentation.}

\label{table:sota_val}
\end{center}
\end{table*}
\renewcommand{\arraystretch}{1}

%% file: sec/4_experiment.tex
\input{table/main_ablation}
\section{Experiments}
\label{main_sec:exp}

\subsection{Experimental Setup}
\noindent{\bf Datasets and metrics.} We conducted the experiments on the nuScenes \cite{nuscenes} and View-of-Delft (VoD) \cite{vod} datasets. 
The nuScenes dataset consists of 1,000 driving scenes, each scene is approximately 20 seconds long with 360-degree coverage from six cameras, five radars, and one LiDAR sensor. Keyframes are annotated at 2Hz, covering 10 object classes. Our evaluation follows the official nuScenes benchmark metrics, including mean Average Precision (mAP) and nuScenes Detection Score (NDS). 

The VoD dataset comprises 8,693 frames, each capturing forward view using a single 4D radar, a stereo camera, and a LiDAR. Evaluation is performed using mAP measured over the Entire Annotated Area (EAA) and within the Driving Corridor (RoI) for three object classes (Car, Pedestrian, Cyclist). Since the VoD test server is not yet available, evaluation is conducted on the validation set.

\noindent{\bf Implementation details.}
For the student model, we utilized ResNet \cite{resnet} and ConvNeXt~\cite{convnext} as the image backbone networks and a modified Pillar Feature Network \cite{crtfusion} building on PointPillars~\cite{pointpillars} as the radar backbone network.
The teacher model employs the SECOND~\cite{SECOND} backbone for processing LiDAR point clouds. Following the SOLOFusion~\cite{solofusion}, we incorporate a streaming-based parallel temporal fusion mechanism that leverages BEV features from the preceding eight frames. During the training, the teacher model weights are frozen while the student model is trained for $60$ epochs. Detailed training configurations and hyperparameter settings are provided in the supplementary material.

\subsection{Comparison to the state-of-the-art}
Table~\ref{table:sota_val} compares RCTDistill with existing camera-radar fusion and camera-only 3D object detectors on both the nuScenes validation and test sets. RCTDistill sets a new state-of-the-art performance across various backbone configurations, consistently surpassing previous methods. With ResNet-50, RCTDistill achieves 62.2\% NDS and 55.2\% mAP while maintaining the highest efficiency at 26.2 FPS. For the ResNet-101 backbone configuration, RCTDistill exhibits significant performance gains of 1.1\% in NDS and 1.9\% in mAP compared to SpaRC \cite{sparc}. Moreover, on the test set, our method outperforms all existing radar-camera fusion models. Even when a single frame input is used (RCTDistill-S), the proposed method achieves superior results. Notably, RCTDistill achieves the highest inference speed (up to 28.0 FPS) across all evaluated configurations.

\begin{figure*}[t]
    \centering
        \includegraphics[width=0.88\linewidth]{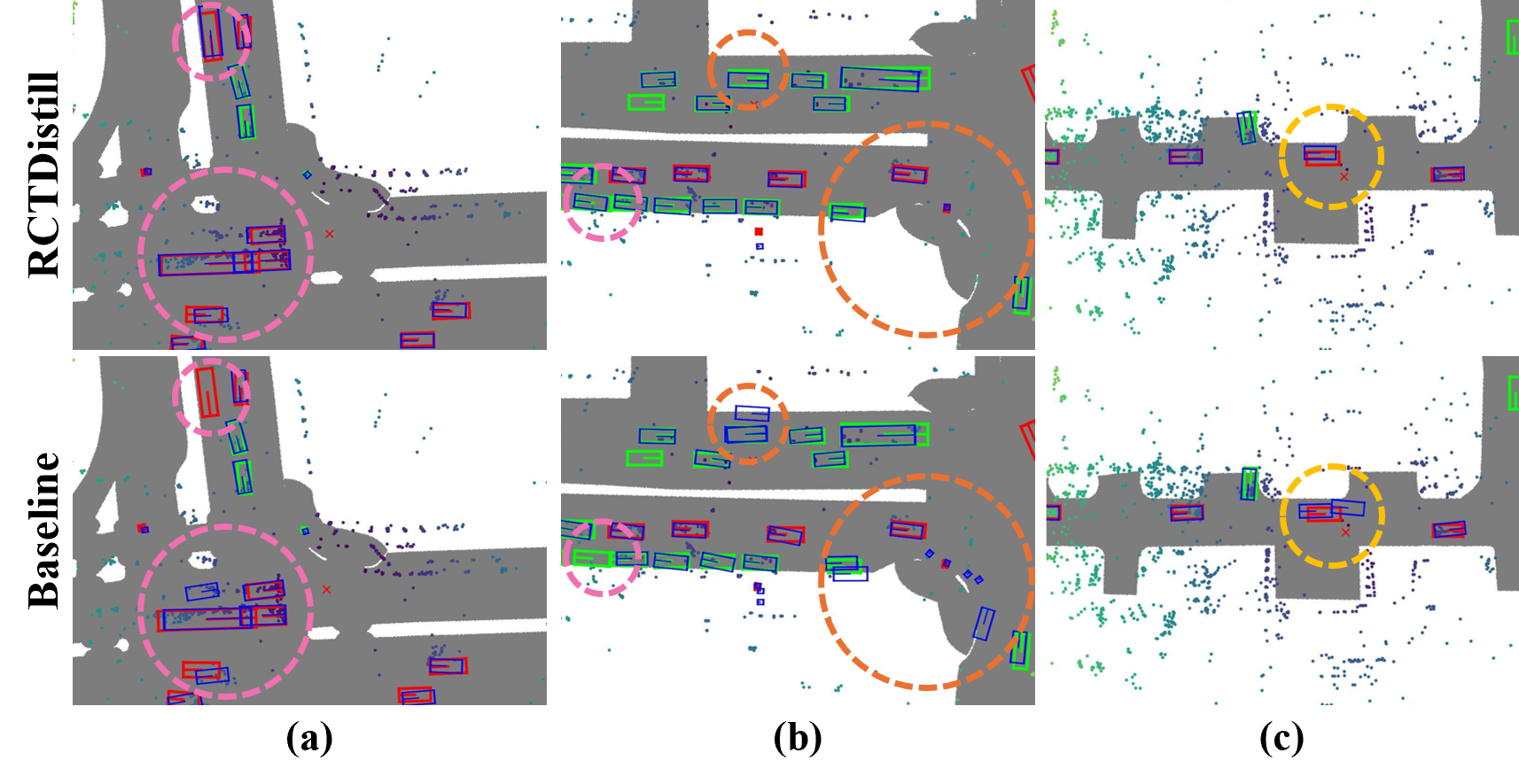}
        \caption{\textbf{Qualitative results comparing RCTDistill and Baseline.} Blue, green, and red boxes denote predictions, static ground truth, and dynamic ground truth, respectively. Pink circles highlight areas where the baseline model fails to detect objects, but RCTDistill successfully identifies them. Orange and yellow circles indicate regions where false positives have been corrected.}
    \label{fig:qualitative_results}
\end{figure*}

\subsection{Ablation Studies}
We conducted ablation studies on both the nuScenes and VoD validation sets.
All experiments were conducted using a ResNet-50 image backbone and trained for 20 epochs.

\input{table/region_ablation}
\input{table/4dRadar}

\noindent{\bf Components analysis.} 
Table~\ref{table:ablation_main} presents an ablation study of the RCTDistill, focusing on the contributions of RAKD, RDKD, and TKD. Each component, when applied individually, improves performance by at least 1.4\% in mAP and 1.8\% in NDS. Pairwise combinations further boost results, while integrating all three components achieves the highest performance, reaching 58.4\% NDS and 51.0\% mAP. This demonstrates the effective synergy between spatial, temporal, and relational distillation techniques.

\noindent{\bf Impact of KD region selection on model performance.}
Table~\ref{table:combined_ablation} presents an ablation study evaluating the performance impact of applying KD to different regions for each module. 
In Table~\ref{table:MSFD_ablation}, we compare our proposed elliptical Gaussian mask with previous methods that adopt different strategies for determining knowledge distillation regions. BEVDistill \cite{bevdistill} employs a circular Gaussian distribution centered on the object. In contrast, CRKD \cite{crkd} proposes MSFD, which expands the ground-truth bounding box region based on object velocity and range thresholds.
We replaced RAKD with these methods in our RCTDistill framework for comparison. While the previous approaches outperform the baseline, our method achieves greater performance gains—specifically, a 2.2\% improvement in both NDS and mAP. This improvement may be attributed to the explicit modeling of sensor-specific uncertainty.

Table~\ref{table:TKD_ablation} demonstrates the effectiveness of our proposed TKD method compared to existing temporal KD approaches. While STXD~\cite{stxd} relies on constructing temporal similarity maps between past and current BEV features for KD, and VCD~\cite{vcd} applies distillation over circular Gaussian masks across all time steps, our method employs a simpler yet more effective approach. TKD leverages only current-frame ground truth boxes and velocity information to identify adaptive distillation regions that reflect potential vehicle movement patterns. Our method outperforms existing temporal KD approaches, achieving significant improvements of 2.0\% in NDS and 1.4\% in mAP over the baseline.  Additional analysis is provided in the Supplementary Material.

Table~\ref{table:RDKD_ablation} highlights the effectiveness of our RDKD compared to conventional approaches that apply relation KD across the entire region. MonoDistill \cite{monodistill} proposes distillation between feature maps, and CRKD \cite{crkd} presents a ReID strategy that utilizes distillation on affinity maps. In contrast, our method selectively targets specific regions. Our approach outperforms the existing methods, improving mAP by at least 0.9\% and NDS by 0.8\%. 

\noindent{\bf Performance comparison on VoD dataset.}
Table~\ref{table:ablation_vod} presents a performance comparison between RCTDistill and existing models based on the VoD \cite{vod} validation set. RCTDistill achieves state-of-the-art performance by outperforming the latest SGDet3D method \cite{sgdet3d}, achieving a 2.62\% increase in EAA AP and a 4.83\% improvement in RoI AP. These results demonstrate that RCTDistill consistently delivers superior performance across both the nuScenes \cite{nuscenes} and VoD \cite{vod} datasets, underscoring its robust performance across diverse environments and sensor characteristics.

\noindent{\bf Qualitative results.} 
Figure~\ref{fig:qualitative_results} presents a qualitative comparison between the proposed RCTDistill and the baseline model.
The highlighted circles illustrate RCTDistill's enhanced ability to detect missed objects (pink),  reduce false positives caused by range and azimuth uncertainties (orange), and resolve false positives caused by dynamic object movement (yellow).
These results demonstrate its effectiveness in addressing the limitations of previous methods.

%% file: table/main_ablation.tex
\setlength{\tabcolsep}{1.1mm}  
\renewcommand{\arraystretch}{1.1}  
\begin{table}[ht!]
\begin{center}
\small
\begin{tabular}{c|ccc|cc}  
\Xhline{1.2pt}  
\textbf{Methods} & \textbf{RAKD} & \textbf{RDKD}  & \textbf{TKD} & \textbf{NDS}$\uparrow$ & \textbf{mAP}$\uparrow$\\ 
\hline\noalign{\hrule height 0.3pt}  
Baseline   &  &  &  & 55.0 & 47.1 \\
\hline\noalign{\hrule height 0pt}  
\multirow{8}{*}{\begin{tabular}[c]{@{}c@{}}RCTDistill\end{tabular}} & \checkmark &  &   & 57.2 & 49.3 \\
&  & \checkmark &  & 56.8 & 49.0 \\
&  &  & \checkmark & 57.0 & 48.5 \\
\Xcline{2-6}{0.1pt}
&\checkmark  &  & \checkmark & 57.5 & 50.1 \\
&  & \checkmark & \checkmark &  57.4 & 49.3 \\
&\checkmark  & \checkmark &  &   57.6 & 49.9 \\
\Xcline{2-6}{0.1pt}
& \checkmark & \checkmark & \checkmark & 58.4 & 51.0 \\

\Xhline{1.2pt}  
\end{tabular}
\caption{Ablation study for evaluating the main components.}
\label{table:ablation_main} 
\end{center}
\end{table}

%% file: table/region_ablation.tex
\setlength{\tabcolsep}{4pt} 
\renewcommand{\arraystretch}{1.1} 

\begin{table}[htbp]
\centering
\definecolor{lightblue}{RGB}{220, 230, 255}

\begin{subtable}[t]{0.43\textwidth}
\centering
\resizebox{\columnwidth}{!}{
\begin{tabular}{c|c|c|c|cc}
\Xhline{1.2pt}
\textbf{Methods} & \textbf{BEVDistill} \cite{bevdistill} & \textbf{CRKD} \cite{crkd} & \textbf{Ours} & \textbf{NDS$\uparrow$} & \textbf{mAP$\uparrow$} \\
\hline
Baseline & & & & 55.0 & 47.1 \\
\Xhline{0.1pt}
\multirow{3}{*}{RAKD} & \checkmark & & & 56.4 & 48.3 \\
 &  &  \checkmark & & 56.7 & 48.6 \\
 &  &  & \checkmark & 57.2 & 49.3 \\
\Xhline{1.2pt}
\end{tabular}
}
\caption{Performance Comparison Based on Mask Types.}
\label{table:MSFD_ablation}
\end{subtable}

\hspace{9cm}

\begin{subtable}[t]{0.43\textwidth}
\centering
\resizebox{\columnwidth}{!}{
\begin{tabular}{c|c|c|c|cc}
\Xhline{1.2pt}
\textbf{Methods} & \textbf{VCD} \cite{vcd} & \textbf{STXD} \cite{stxd} & \textbf{Ours} & \textbf{NDS$\uparrow$} & \textbf{mAP$\uparrow$} \\ 
\hline
Baseline & & & & 55.0 & 47.1 \\
\Xhline{0.1pt}
\multirow{3}{*}{TKD} & \checkmark & & & 55.9 & 48.3 \\
 & & \checkmark & & 56.5 & 48.1 \\
 & & & \checkmark & 57.0 & 48.5 \\
\Xhline{1.2pt}
\end{tabular}
}
\caption{Performance Comparison Based on Temporal Knowledge Distillation Methods.}
\label{table:TKD_ablation}
\end{subtable}

\hspace{9cm}

\begin{subtable}[t]{0.43\textwidth}
\centering
\resizebox{\columnwidth}{!}{
\begin{tabular}{c|c|c|c|cc}
\Xhline{1.2pt}
\textbf{Methods} & \textbf{MonoDistill} \cite{monodistill} & \textbf{CRKD} \cite{crkd} & \textbf{Ours} & \textbf{NDS$\uparrow$} & \textbf{mAP$\uparrow$} \\ 
\hline
Baseline & & & & 55.0 & 47.1 \\
\Xhline{0.1pt}
\multirow{3}{*}{RDKD} & \checkmark & & & 56.0 & 47.8 \\
 &  &  \checkmark & & 55.9 & 48.1 \\
 &  & & \checkmark & 56.8 & 49.0 \\
\Xhline{1.2pt}
\end{tabular}
}
\caption{Performance Comparison by Relation KD Application Scope.}
\label{table:RDKD_ablation}
\end{subtable}

\caption{Ablation studies of proposed KD methods.}
\label{table:combined_ablation}
\end{table}

%% file: table/4dRadar.tex
\setlength{\tabcolsep}{1.1mm}  
\renewcommand{\arraystretch}{0.95}  

\begin{table}[t!]
\begin{center}
\small
\definecolor{lightblue}{RGB}{220, 230, 255}
\begin{tabular}{c|c|c|c}  
\Xhline{1.2pt}  
\textbf{Methods} & \textbf{Input} & \textbf{EAA AP}$\uparrow$ & \textbf{RoI AP}$\uparrow$\\ 
\hline\noalign{\hrule height 0.3pt}  
BEVFusion \cite{bevfusion} & C+R & 49.25 & 68.52 \\
LXL \cite{lxl} & C+R & 56.31 & 72.93 \\
RCBEVDet \cite{rcbevdet} & C+R & 49.99 & 69.80 \\
HGSFusion \cite{gu2024hgsfusion} & C+R & 58.96 & 79.46 \\
SGDet3D \cite{sgdet3d} & C+R & 59.75 & 77.42 \\
\rowcolor{lightblue}
RCTDistill & C+R & \bf{62.37} & \bf{82.25}\\

\Xhline{1.2pt}  
\end{tabular}
\caption{Performance Comparison on VoD \cite{vod} Dataset. RoI denotes the driving corridor area. EAA: Entire Annotated Area.}
\label{table:ablation_vod} 
\end{center}
\end{table}

%% file: sec/5_conclusion.tex
\section{Conclusions}
\label{main_sec:con}
In this paper, we presented RCTDistill, a novel cross-modal knowledge distillation framework designed to improve temporal radar-camera fusion for 3D object detection. RCTDistill effectively addresses key challenges, including sensor-specific uncertainty and temporal misalignment, leading to significant performance gains. The proposed RAKD module mitigates sensor-specific uncertainty by applying elliptical Gaussian regions, thereby enhancing the quality of BEV representations. TKD alleviates temporal misalignment caused by dynamic objects by aligning historical BEV features more accurately with current LiDAR features. Lastly, RDKD improves feature discrimination by transferring relational knowledge from the teacher model, enabling the student model to more effectively distinguish foreground from background features. Overall, RCTDistill establishes a new state-of-the-art in radar-camera-based 3D object detection.

%% file: sec/6_acknowledgement.tex
\section{Acknowledgement}
This work was supported by Institute of Information \& communications Technology Planning \& Evaluation (IITP) grant funded by the Korea government(MSIT) [NO. RS-2021-II211343, Artificial Intelligence Graduate School Program (Seoul National University)], the National Research Foundation (NRF) funded by the Korean government (MSIT) (No. RS-2024-00421129), and the Technology Innovation Program (20018112, Development of autonomous manipulation and gripping technology using imitation learning based on visualtactile sensing) funded By the Ministry of Trade, Industry \& Energy(MOTIE, Korea).